\documentclass[11pt]{article}

\usepackage[preprint]{acl}

\usepackage{times}
\usepackage{latexsym}
\usepackage{lipsum}
\usepackage[T1]{fontenc}

\usepackage[utf8]{inputenc}

\usepackage{graphicx}
\graphicspath{ {./images/} }

\usepackage{xcolor} 
\definecolor{LightGray}{gray}{0.9}

\usepackage{microtype}
\usepackage{inconsolata}

\usepackage{multirow}

\title{LastResort at SemEval-2024 Task 3: Exploring Multimodal Emotion Cause Pair Extraction as Sequence Labelling Task}

\author{Suyash Vardhan Mathur\thanks{Equal contribution.} \\
  IIIT Hyderabad \\
  {\footnotesize \texttt{suyash.mathur@research.iiit.ac.in\ } } \\\And
  Akshett Rai Jindal\footnotemark[1] \\
  IIIT Hyderabad \\
  {\footnotesize \texttt{\ akshett.jindal@research.iiit.ac.in\ } } \\\AND
  Hardik Mittal \\
  IIIT Hyderabad \\
  {\footnotesize \texttt{\ hardik.mittal@research.iiit.ac.in\ } } \\\And
  Manish Shrivastava \\
  IIIT Hyderabad \\
  {\footnotesize \texttt{\ m.shrivastava@iiit.ac.in} } }

\begin{document}
\maketitle
\begin{abstract}
Conversation is the most natural form of human communication, where each utterance can range over a variety of possible emotions. While significant work has been done towards the detection of emotions in text, relatively little work has been done towards finding the cause of the said emotions, especially in multimodal settings. SemEval 2024 introduces the task of Multimodal Emotion Cause Analysis in Conversations, which aims to extract emotions reflected in individual utterances in a conversation involving multiple modalities (textual, audio, and visual modalities) along with the corresponding utterances that were the cause for the emotion. In this paper, we propose models that tackle this task as an utterance labeling and a sequence labeling problem and perform a comparative study of these models, involving baselines using different encoders, using BiLSTM for adding contextual information of the conversation, and finally adding a CRF layer to try to model the inter-dependencies between adjacent utterances more effectively. In the official leaderboard for the task, our architecture was ranked 8\textsuperscript{th}, achieving an F1-score of 0.1759 on the leaderboard. We also release our code here\footnote{\url{github.com/akshettrj/semeval2024_task03}}.
\end{abstract}

\section{Introduction}
Emotion Analysis is one of the fundamental and earliest sub-fields of NLP that focus on identifying and categorizing emotions that are expressed in text. Earlier, research in this domain focused on Emotion Detection in news articles and headlines (\citealp{lei-et-al-2014}; \citealp{abdul-mageed-ungar-2017-emonet}). However, later Emotion Recognition in Conversation gained popularity due to the widespread availability of public conversation data (\citealp{gupta2017sentiment}). Recently, the task of emotion cause analysis has gained traction, which tries to identify the causes behind certain emotions (\citealp{xia2019emotion}). This has widespread application such as building chatbots that can identify the emotions of the user and even identify the cause behind the emotions to perform certain actions (\citealp{pamungkas2019emotionallyaware}). For instance, companies can identify causes behind dissatisfaction in customer interactions and take appropriate measures (\citealp{yun2022effects}), AI-driven therapeutic insights can be gained using such models (\citealp{d2020ai}), social media content moderation can be better done (\citealp{sawhney2021towards}), work management and team management by companies can be improved (\citealp{benke2020chatbot}).

In the task \citealp{ECAC2024SemEval}, we tackle the problem of Multimodal Emotion Cause Pair Extraction, where given a set of utterances in a conversation, we must identify the following:

\textbf{1. Emotion} of every utterance (if any). These emotions can be one of Ekman's six basic emotions (\citealp{ekman1999basic}).

\textbf{2. Cause} of these emotions, which is considered as the utterance that explicitly expresses an event or argument that is highly linked to the corresponding emotion.

Our proposed system tackles the task in a 3-step fashion -- (a) First, we train a model to identify the emotions that are expressed in individual utterances in a conversation. (b) Next, we train a model to identify whether an utterance can be a cause of an emotion expressed in another/same utterance (candidate causes). (c) Finally, we train a model to pair emotion-utterances with their causes among the possible candidate causes. For both the (a) and (b) models we experiment with 3 basic architectures -- (i) a simple Neural Network to determine the class of emotion (N-class classifier) and another Neural Network to identify whether the utterance is a candidate cause or not (binary classification). (ii) A BiLSTM (\citealp{sak2014long}) architecture that accounts for the surrounding context of the conversation while doing the N-class and binary-classification. (iii) A BiLSTM CRF (\citealp{lafferty2001conditional}) architecture which accounts for the surrounding emotions as well while doing the N-class classification. We also experiment with different encoders for the three modalities.




\section{Background}

\subsection{Dataset}

The dataset used for this problem is \textbf{Emotion-Cause-in-Friends} prepared by \citealp{wang2023multimodal} specifically for this task. It has been prepared using conversations from the popular 1994 sitcom \textit{Friends} as the source. This dataset contains 1,344 conversations made up of a total of 13,509 utterances, each conversation containing an average of 10 utterances. For each utterance, the dataset has an annotated transcript (covering text modality) and the corresponding video clip (covering visual and auditory modalities) from the show.

\begin{figure}
    \centering
    \includegraphics[scale=0.4]{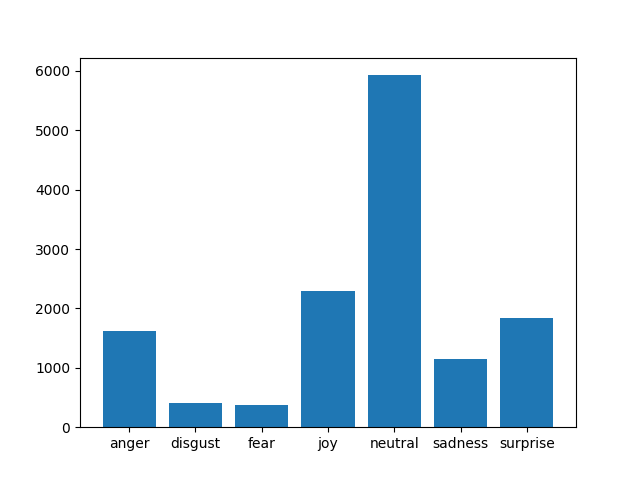}
    \caption{Emotion frequency in the dataset}
    \label{fig:emotion-frequencies}
\end{figure}

Each utterance is annotated with the emotion depicted by it, which is one of: \texttt{anger}, \texttt{disgust}, \texttt{fear}, \texttt{joy}, \texttt{neutral}, \texttt{sadness} and \texttt{surprise}. The dataset is highly skewed in terms of the frequency of different emotions in the dataset (see Figure~\ref{fig:emotion-frequencies}). 
Further, the emotion-causes pairs for all the non-\texttt{neutral} utterances are provided in the dataset in a separate list.

\begin{figure*}
    \centering
    \includegraphics[scale=0.2]{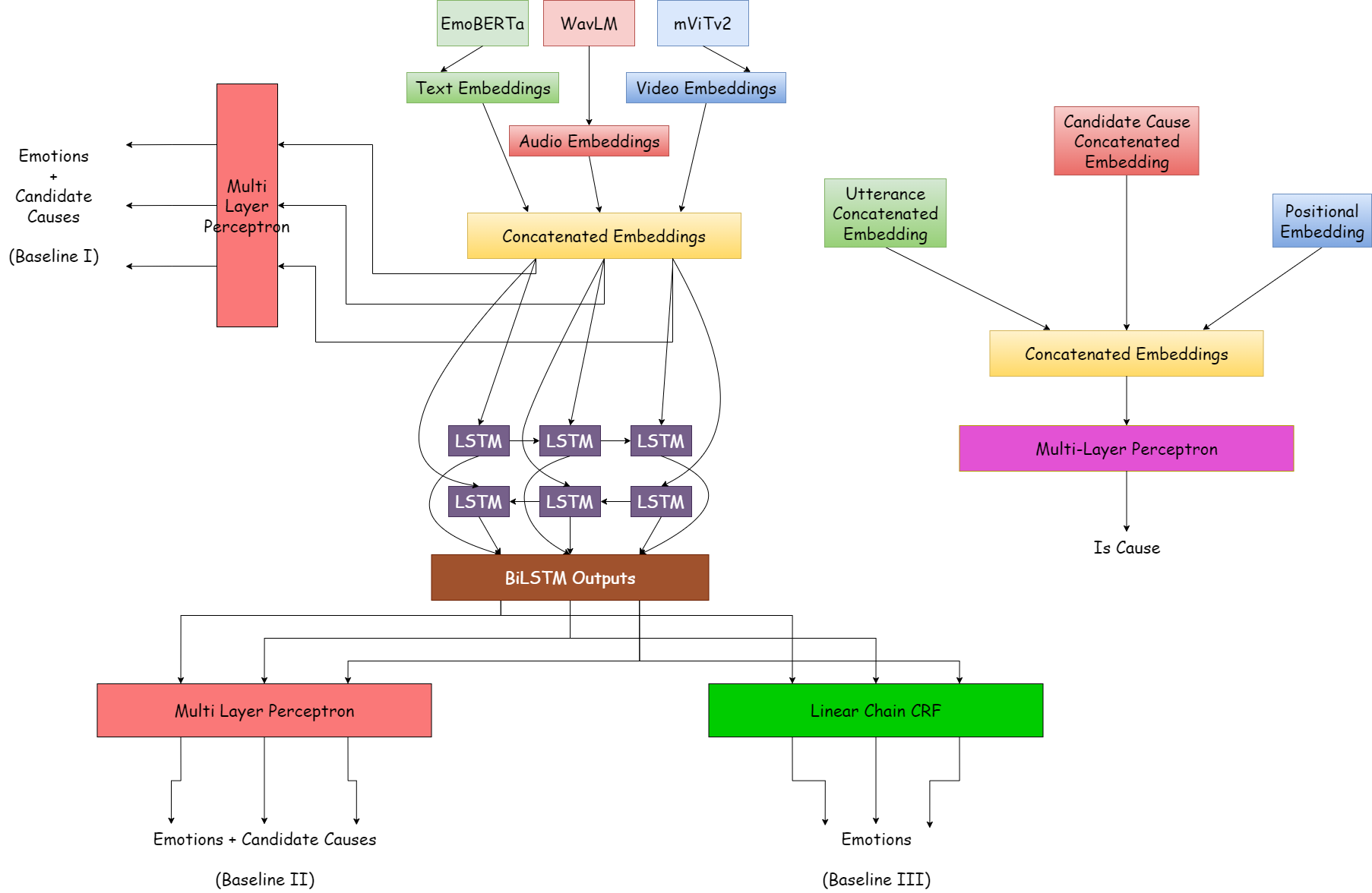}
    \caption{Model Architecture}
    \label{fig:model-architecture}
\end{figure*}

The task MC-ECPE expects the model to take a list of such conversations and predict the emotion and emotion-cause pairs labels.

\subsection{Related Work}

A lot of work has been done in the field of emotion analysis in textual settings. Soon, work began on extracting not only the emotion but also the cause of that extracted emotion. People employed mainly two approaches for emotion cause analysis - 1. Extracting the potential causes given an emotion (\citealp{lee-etal-2010-text}; \citealp{chen-etal-2010-emotion}; \citealp{gui-et-al-2016b}) and 2. Extracting the emotion-cause pairs jointly (\citealp{xia-ding-2019-emotion}; \citealp{ding-etal-2020-ecpe}; \citealp{wei-etal-2020-effective}).

\citealp{poria-et-al-2020} was the first to introduce the task of extracting emotion-cause in conversations but their focus was also only on the textual dialogues. However, in our natural way of conversation, we rely on things like facial expressions, voice intonations for determining the emotion of the speaker. We also rely on auditory and visual scenes to determine the cause of the speaker's emotions. Hence, it is clear that \textbf{Emotion-Cause Pair Extraction (ECPE)} is a multimodal task requiring at least three modalities: \textbf{text, audio} and \textbf{video}. \citealp{Busso2008IEMOCAPIE}; \citealp{mckeown-el-al-2012}; \citealp{wei-et-al-2022} and \citealp{poria-etal-2019-meld} worked in the field of multimodal emotion analysis in conversations but they did not consider the emotion causes.

The task of MC-ECPE was first worked on by \citealp{ffwang-et-al-2021}.




\section{System Overview}

\subsection{Baseline I: Utterance labeling}
Our baseline model treats the problem as a simple \textbf{utterance labeling task}. We use pre-trained text, audio, and image encoders to encode the individual modalities and use these to train three models that can identify the emotions in the utterances, the candidate cause utterances, and finally identify valid emotion and cause utterance(s) pairs.
\begin{itemize}
    \item \textbf{Text Encoding:} For encoding the transcription of each utterance, we use pre-trained BERT (\citealp{devlin2018bert}) embeddings as the baseline embeddings. Additionally, we finetune DeBERTa-Base (\citealp{he2020deberta}) on the training data for our experiments. DeBERTa makes use of a disentangled attention mechanism and an enhanced masked encoder to improve upon BERT's performance in a variety of tasks. Finally, we also tried RoBERTa-Large and (\citealp{liu2019roberta}) pre-trained EmotionRoBERTa-Base\footnote{\url{https://huggingface.co/SamLowe/roberta-base-go\_emotions}} which is publicly available RoBERTa-base model finetuned on the Go Emotions dataset (\citealp{demszky-etal-2020-goemotions}). For every text encoder, we perform mean-pooling of the word embeddings to get the textual representation of the utterance.
    \item \textbf{Video Encodings:} For encoding the videos, we sampled 16 equally spaced frames from the video and mean-pooled the embeddings for the 16 frames. For encoding these 16 images, we used MViTv2-small (\citealp{li2022mvitv2}) encoder, which achieves state-of-the-art performance on the Kinetics video detection task (\citealp{kay2017kinetics}), which makes it an obvious choice for recognizing activities happening in the conversations relevant for emotion/cause detection.
    
    \item \textbf{Audio Encodings:} We used WavLM (\citealp{chen2022wavlm}) for generating audio embeddings, which is trained on large audio data using masked speech representation and denoising in pre-training, making it suitable for various downstream speech tasks. We also try Wav2Vec2-Large (\citealp{baevski2020wav2vec}), which is trained by masking speech input in latent space and solving a contrastive task defined over a quantization of the latent representations which are jointly learned.
\end{itemize}
The model architecture is a combination of \textbf{three steps}, each of which is described below: 

\subsection*{Step 1 -- Emotion Classification} First, we concatenate the text, audio and video embeddings from the respective encoders and pass these concatenated embeddings into a dense layer, on which a Softmax function is applied to get the probability distribution over 7 classes (6 emotions and one neutral class). Due to a skewed distribution of the emotion labels in the dataset, we make use of \textbf{weighted Cross Entropy loss} to train the model, where the weights are taken as inverse of the frequency of the labels in the training dataset.

\subsection*{Step 2 -- Candidate Cause Identification} For identifying the candidate cause, we similarly pass concatenated embeddings through a dense layer with a Sigmoid function, which predicts the probability of whether the utterance is a candidate cause or not. Binary Cross Entropy Loss is used to train the model.

\subsection*{Step 3 -- Emotion-Cause pairing} For pairing the emotion utterances with the candidate causes, we concatenate the represenations for the emotion utterance and the cause utterance, with a distance embedding. This distance embedding is generated by giving positional embedding to each utterance, sampled from a Normal Distribution. This representation is passed through a dense layer with a Sigmoid function, which learns to predict the probability of the emotion-cause utterance pair being a valid emotion-cause pair or not for the given conversation, trained using Binary Cross Entropy Loss.

\subsection{Baseline II: BiLSTM Architecture}

The BiLSTM architecture is inspired by the work in \citealp{ffwang-et-al-2021}. While the Baseline I architecture treats the emotion and cause classification independently for each utterance, it is dependent on the surrounding context of the conversation too. Thus, the BiLSTM architecture models the problem as a \textbf{Sequence Labeling task}. We use the best encoders in the Baseline I architecture for generating the embeddings in this architecture.

\subsection*{Step 1 -- Emotion Classification}
Similar to the Baseline Model, we concatenate the embeddings of the three modalities, and pass them to a stacked BiLSTM. On top of the BiLSTM outputs, we apply a 7-class classifier to obtain the emotion category distribution. Similar to the Baseline I, weighted cross-entropy loss is used.

\subsection*{Step 2 -- Candidate Cause Identification}
For Candidate Cause prediction, similarly, the concatenated embeddings are passed through a BiLSTM on top of which a binary classifier is applied. 

\subsection*{Step 3 -- Emotion-Cause Pairing}
The Emotion-Cause pairing model remains the same in this architecture as the Baseline I model.

\vspace{1em}
In this architecture, BiLSTM provides the advantages of bidirectional and longer contexts which should help understand the emotions present in utterances better. This is because in a conversation, it is possible that the emotions are not just dependent on the current utterance, but on surrounding multimodal utterances as well.

\subsection{Baseline III: BiLSTM-CRF Architecture}

In the BiLSTM model, each classification decision was conditionally independent. Linear-chain CRFs are models generally used to model structured data where one output influences its neighboring outputs as it models the various transition probabilities, and have been extensively used with BiLSTMs for sequence labeling (\citealp{huang2015bidirectional}). This could be useful for emotion predictions because the emotion of one utterance is generally influenced by the emotions in its previous utterances. For instance, an utterance with happiness generally tends to be followed by another happiness utterance.

\subsection*{Step 1 -- Emotion Classification}
For this architecture, we add a CRF layer on top of the BiLSTM layers, and make use of the CRF-loss to train the model instead of Cross-Entropy loss as in the previous architectures. This loss models the transitions between the labels in the architecture, modelling the task as a more complex sequence labeling task. Thus, while the BiLSTM layer learns more about the language and emotions expressed through the language, the CRF layer tries to learn about the relations between the emotions. 

\subsection*{Step 2 -- Candidate Cause Identification}
For Candidate Cause prediction, the architecture remains the same as in Baseline II. This is because the transitions between cause labels (being cause of an emotion in an utterance or not) does not make intuitive sense, and using BiLSTMs to capture surrounding context from other utterances is what seems more appropriate.

\subsection*{Step 3 -- Emotion-Cause Pairing}
The Emotion-Cause pairing model remains the same in this architecture as the Baseline I \& II models.

\section{Experimental Setup}
\begin{table*}
    \centering
    \footnotesize
    \begin{tabular}{|p{10em}|p{1.25em}p{1.25em}p{1.25em} |p{1.25em}p{1.25em}p{1.25em}|p{1.25em}p{1.25em}p{1.25em}|p{2em}p{2em}|p{2em}p{2em}|}
        \hline
        \multirow{2}{*}{\textbf{\footnotesize Model Name}} & \multicolumn{3}{|c|}{\textbf{\footnotesize Emotion}} & \multicolumn{3}{|c|}{\textbf{\footnotesize Candidate Cause}} & \multicolumn{3}{|c|}{\textbf{\footnotesize Emotion-Cause}} & \multicolumn{2}{|c|}{{\textbf{\footnotesize Emotion-Cause}}} & \multicolumn{2}{|c|}{\multirow{2}{*}{\textbf{\footnotesize Leaderboard}}} \\
        & \multicolumn{3}{|c|}{\textbf{\footnotesize Detection}} & \multicolumn{3}{|c|}{\textbf{\footnotesize Detection}} & \multicolumn{3}{|c|}{\textbf{\footnotesize Pairing}} & \multicolumn{2}{|c|}{\textbf{\footnotesize Pairing (Eval.)}} & \\
        \hline
        & \textbf{\footnotesize P} & \textbf{\footnotesize R} & \textbf{\footnotesize F1} & \textbf{\footnotesize P} & \textbf{\footnotesize R} & \textbf{\footnotesize F1} & \textbf{\footnotesize P} & \textbf{\footnotesize R} & \textbf{\footnotesize F1} & \textbf{\footnotesize wt. F1} & \textbf{\footnotesize Macro F1} & \textbf{\footnotesize wt. F1} & \textbf{\footnotesize Macro F1} \\
        \hline
        {\footnotesize \textbf{Baseline I}} & {\footnotesize } & {\footnotesize } & {\footnotesize } & {\footnotesize } & {\footnotesize } & {\footnotesize } & {\footnotesize } & {\footnotesize } & {\footnotesize } & {\footnotesize } & {\footnotesize } \\


        {\footnotesize BERT + WavLM + MViTv2} & {\footnotesize 0.61} & {\footnotesize 0.52} & {\footnotesize \textbf{0.55}} & {\footnotesize 0.71} & {\footnotesize 0.71} & {\footnotesize \textbf{0.71}} & {\footnotesize 0.93} & {\footnotesize 0.87} & {\footnotesize \textbf{0.89}} & {\footnotesize \textbf{0.26}} & {\footnotesize 0.20} & {\footnotesize 0.182} & {\footnotesize 0.165} \\
        
        \hline
        
        {\footnotesize EmotionRoBERTa + WavLM + MViTv2} & {\footnotesize 0.55} & {\footnotesize 0.45} & {\footnotesize 0.47} & {\footnotesize 0.67} & {\footnotesize 0.66} & {\footnotesize 0.66} & {\footnotesize 0.93} & {\footnotesize 0.86} & {\footnotesize \textbf{0.89}} & {\footnotesize 0.20} & {\footnotesize 0.18} & {\footnotesize \textbf{0.187}} & {\footnotesize 0.170} \\

        \hline
        
        {\footnotesize DeBERTa (finet.) + WavLM + MViTv2} & {\footnotesize 0.44} & {\footnotesize 0.36} & {\footnotesize 0.38} & {\footnotesize 0.60} & {\footnotesize 0.60} & {\footnotesize 0.60} & {\footnotesize 0.92} & {\footnotesize 0.85} & {\footnotesize 0.87} & {\footnotesize 0.10} & {\footnotesize 0.10} & {\footnotesize 0.094} & {\footnotesize 0.094} \\
        
        \hline
        
        {\footnotesize RoBERTa-L + WavLM + MViTv2} & {\footnotesize 0.59} & {\footnotesize 0.47} & {\footnotesize 0.49} & {\footnotesize 0.66} & {\footnotesize 0.65} & {\footnotesize 0.66} & {\footnotesize 0.93} & {\footnotesize 0.86} & {\footnotesize 0.88} & {\footnotesize 0.21} & {\footnotesize 0.19} & {\footnotesize 0.180} & {\footnotesize 0.165} \\
        
        \hline
        
        {\footnotesize EmotionRoBERTa + Wav2Vec2 + MViTv2} & {\footnotesize 0.55} & {\footnotesize 0.47} & {\footnotesize 0.48} & {\footnotesize 0.67} & {\footnotesize 0.67} & {\footnotesize 0.67} & {\footnotesize 0.93} & {\footnotesize 0.87} & {\footnotesize \textbf{0.89}} & {\footnotesize 0.21} & {\footnotesize 0.20} & {\footnotesize 0.172} & {\footnotesize 0.170}  \\
        \hline
        {\footnotesize BiLSTM \textbf{(Baseline II)}} & {\footnotesize 0.55} & {\footnotesize 0.51} & {\footnotesize 0.52} & {\footnotesize 0.67} & {\footnotesize 0.67} & {\footnotesize 0.67} & {\footnotesize 0.93} & {\footnotesize 0.86} & {\footnotesize \textbf{0.89}} & {\footnotesize 0.22} & {\footnotesize \textbf{0.21}} & {\footnotesize 0.184} & {\footnotesize \textbf{0.179}}  \\
        \hline
        {\footnotesize BiLSTM + CRF\textbf{(Baseline III)}} & {\footnotesize 0.53} & {\footnotesize 0.56} & {\footnotesize 0.54} & {\footnotesize 0.67} & {\footnotesize 0.67} & {\footnotesize 0.67} & {\footnotesize 0.93} & {\footnotesize 0.86} & {\footnotesize \textbf{0.89}} & {\footnotesize 0.24} & {\footnotesize 0.18} & {\footnotesize 0.165} & {\footnotesize 0.172}  \\
        
        \hline
    \end{tabular}
    \caption{Results for baselines on the ECAC dataset}
    \label{table:results}
\end{table*}

We perform a random shuffle and use a 90-10\% split for the train-validation split. The test set was provided by the authors, but its gold labels have not been made public.

The experiments involving Baseline II and III use \textit{EmotionRoBERTa + WavLM + MViTv2} configuration. All the experiments involve applying a dropout of 0.3 on the audio, visual and textual embeddings before they are passed on to the main architectures. The BiLSTM for emotion detection consists of 4 layers while the one for candidate cause identification contains 3 layers. The dropout between the stacked layers of the BiLSTM is kept 0.3 as well. We use AdamW optimizer for all the three models, and use a linear learning rate scheduler with warmup for training the models. The Emotion Classification model is trained for 60 epochs, the Candidate Cause Identification model is trained for 40 epochs, and the Emotion-Cause Pairing Model is trained for 40 epochs as well.

In order to train the Emotion-Cause pairing model, we create positive and negative pairs during training. However, while the number of positive pairs is of the order \textit{N}, the number of negative pairs comes to the order of \textit{N\textsuperscript{2}}, and thus we perform a random sampling of the negative pairs to keep the positive and negative samples in the ratio 1:5. This helps us to maintain balance between the positive and negative classes.

%
%
\subsection*{Evaluation Metrics}
We evaluate the 3 steps separately as well apart from evaluating the performance for the final Emotion-Cause pairs:\\
\textbf{Emotion Identification:} We use Weighted Precision, Recall and F1-score for the distribution between the 7 classes (6 emotions and neutral class).\\
\textbf{Candidate Cause Identification:} We again use Weighted Precision, Recall and F1-score for evaluating the prediction between the binary classes -- \textit{is\_candidate\_cause} and \textit{not\_candidate\_cause}.\\
\textbf{Emotion-Cause Pairing:} For evaluating this, we generate positive and negative pairs and use Weighted Precision, Recall and F1-score for evaluating the classification between the positive and negative classes.\\
\textbf{Emotion-Cause Pairs:} Weighted F1-score and Macro F1-score are the official metric used for the final evaluation for the task.

\section{Results and Analysis}
The performance of the three Baselines can be seen in Table \ref{table:results}. During the Evaluation phase, our best ranked submission of Baseline II had Wt. F1 score of 0.1836 and Macro F1 score of	0.1759, ranking 8\textsuperscript{th} on the leaderboard.
\subsection*{Baseline I}
Among the encoders in Baseline I, \textit{BERT + WavLM + MViTv2} configuration performs the best on the validation set, including the individual steps as well as the final emotion-cause pair predictions. However, on the leaderboard, \textit{EmotionRoBERTa + WavLM + MViTv2} gives the best performance, although the difference in the leaderboard scores is marginal among the encoders. This observation might indicate that the test data is a bit different in nature from the training data. 

Better performance of EmotionRoBERTa can be attributed to the fact that the model's weights have already been finetuned towards emotion-related tasks. Further, it seems that finetuning DeBERTa on the training data caused it to overfit, leading to worse performance than vanilla BERT/RoBERTa models. RoBERTa-L performed slightly worse than BERT and EmotionRoBERTa.

Finally, WavLM being the newer architecture, as expected performed better than Wav2Vec2. This is because WavLM is more robust than Wav2Vec2 and it is trained in a combination of supervised and self-supervised learning, making its performance much better.

\subsection*{Baseline II}
We use the \textit{EmotionRoBERTa + WavLM + MViTv2} configuration as encoders for the Baseline II architecture. Contrary to expectation, the wt. F1 score on the leaderboard decreased marginally, while the Macro F1 score increased marginally. This is probably because of the nature of the dataset, where the average length of a conversation is as little as 10, which causes the context of the utterance to be rather limited. In such a situation, the additional context from previous utterances doesn't prove helpful to the model, and might even prove to be noise for the model, leading to the results observed.

\subsection*{Baseline III}
In this, we can observe a significant fall in wt. F1 and slight fall in Macro F1 score from the Baseline I and II architectures. This is in line with the observation of Baseline II that due to the nature of the dataset, sequence labeling it is not necessarily the best way to model it. Further, due to small number of utterances in the conversations, it is likely that the transition between labels needed for CRF doesn't get trained that well and leads to poorer performance.


\section{Conclusion}
In conclusion, we observe that the utterance labeling systems perform as good as sequence labeling systems for this specific dataset. Further, we also see that encoders which are trained on other emotion-related tasks tend to perform better on similar emotion-related tasks.

In future, it is possible to learn joint embeddings over the 3 modalities, which should provide better representations for each utterance (\citealp{Girdhar_2023_CVPR}). Further, it can be experimented to utilize the speaker information for each utterance while creating utterance representations (\citealp{liang2023si}).

\bibliography{custom}

\end{document}